\documentclass[10pt,journal,compsoc]{IEEEtran}

\ifCLASSOPTIONcompsoc
  \usepackage[nocompress]{cite}
\else
  \usepackage{cite}
\fi
\ifCLASSINFOpdf
\else
\fi

\usepackage{amsmath,amssymb,amsfonts}
\usepackage{algorithmic}
\usepackage{graphicx}
\usepackage{textcomp}
\usepackage{xcolor}
\usepackage{siunitx}
\usepackage{float}

\usepackage{makecell}
\usepackage[colorlinks,
            linkcolor=blue,
            anchorcolor=blue,
            citecolor=blue]{hyperref}

\begin{document}
%
\title{Heterogeneous Graph-Based Multimodal Brain Network Learning
}
%
%
%
%

\author{Gen~Shi,
        Yifan~Zhu,
        Wenjin~Liu,
        Quanming~Yao,
        Xuesong Li
\IEEEcompsocitemizethanks{\IEEEcompsocthanksitem G. Shi, X. Li are with the School of Computer Science and Technology, Beijing Institute of Technology, Beijing, China.\protect\\
E-mail: {shigen, lixuesong}@bit.edu.cn
\IEEEcompsocthanksitem Y. Zhu is with the School of Computer Science and Technology, Beijing Institute of Technology, Beijing, China.\protect\\
E-mail: zhuyifan@tsinghua.edu.cn
\IEEEcompsocthanksitem W. Liu is with the Department of Nephrology, Northern Jiangsu People's Hospital, Yangzhou University, Yangzhou, China.\protect\\
E-mail: liuwj1989@hotmail.com
\IEEEcompsocthanksitem Q. Yao is with the Department of Electronic Engineering, Tsinghua University, Beijing, China.\protect\\
E-mail: qyaoaa@tsinghua.edu.cn}
\thanks{Corresponding author: Xuesong Li}
}

\markboth{Journal of \LaTeX\ Class Files,~Vol.~14, No.~8, August~2015}%
{Shell \MakeLowercase{\textit{et al.}}: Bare Advanced Demo of IEEEtran.cls for IEEE Computer Society Journals}
%



\IEEEtitleabstractindextext{%
\begin{abstract}
Graph neural networks (GNNs) provide powerful insights for brain neuroimaging technology from the view of graphical networks.
However, most existing GNN-based models assume that the neuroimaging-produced brain connectome network is a homogeneous graph with single types of nodes and edges. 
In fact, emerging studies have reported and emphasized the significance of heterogeneity among human brain activities, especially between the two cerebral hemispheres. 
Thus, homogeneous-structured brain network-based graph methods are insufficient for modelling complicated cerebral activity states. 
To overcome this problem, in this paper, we present a heterogeneous graph neural network (HebrainGNN) for multimodal brain neuroimaging fusion learning. 
We first model the brain network as a heterogeneous graph with multitype nodes (i.e., left and right hemispheric nodes) and multitype edges (i.e., intra- and interhemispheric edges). 
Then, we propose a self-supervised pretraining strategy based on a heterogeneous brain network to address the potential overfitting problem caused by the conflict between a large parameter size and a small medical data sample size. 
Our results show the superiority of the proposed model over other existing methods in brain-related disease prediction tasks.
Ablation experiments show that our heterogeneous graph-based model attaches more importance to hemispheric connections that may be neglected due to their low strength by previous homogeneous graph models. Other experiments also indicate that our proposed model with a pretraining strategy alleviates the problem of limited labelled data and yields a significant improvement in accuracy. 
\end{abstract}

}

\maketitle
\section{Introduction}
Neuroimaging, as a noninvasive and harmless approach, plays a fundamental role in current neuroscience studies and medical diagnosis.
Today, different types of neuroimaging methods have been implemented to establish accurate imaging for different tissues and different purposes\cite{b1, b2, b3}.
For example, diffusion tensor imaging (DTI) provides information on structural connections via white matter among cerebral regions, while functional magnetic resonance imaging (fMRI) reflects changes in brain temporal activities.
Each type of neuroimaging data provides a unique view of the brain, and the fusion of such multimodal neuroimaging data can play a key role in understanding the development of human cerebral disease and even cognitive processes \cite{b7, b8}.

\begin{figure}[!htbp]
\centerline{\includegraphics[width=\columnwidth]{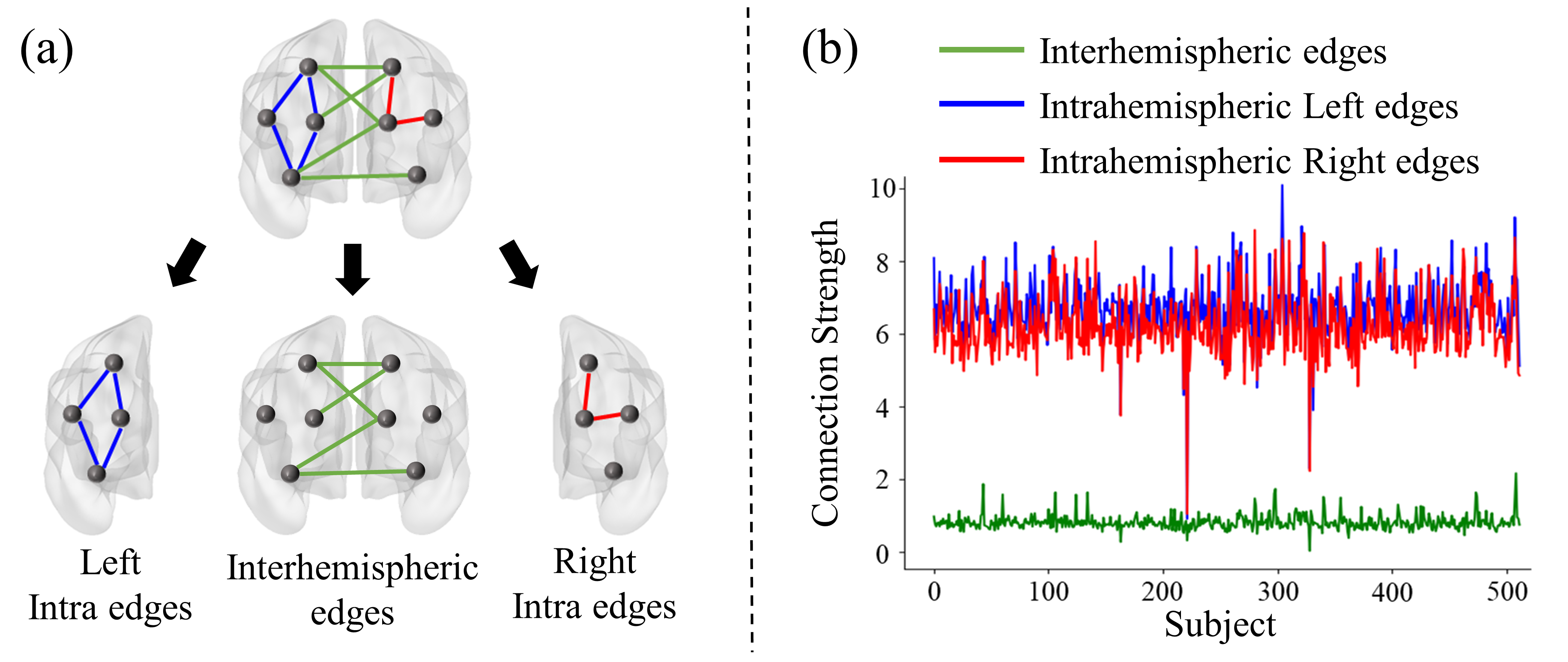}}
\caption{Heterogeneity in the brain network with cerebral hemispheres. (a) Three types of edges, namely, left intrahemispheric edges (blue), interhemispheric edges (green), and right intrahemispheric edges (red). (b) Strength of intra- and interhemispheric edges among subjects in the ADNI dataset. }
\label{fig:DTI_divided}
\end{figure}

Representing brain imaging as a graph is a key approach to model brain activity, with brain regions as nodes and the working mode correlations among different regions as edges. 
Graph-based models such as GNNs have been proposed to integrate both graph structure and node-level feature information to model the brain \cite{b9} and other medical imaging scenarios \cite{b10, b11, b12, b13}. 
However, these methods usually assume that the constructed brain network is a homogeneous graph with only one type of node and edge.
This assumption is far from the reality of the human brain. 
In fact, a large number of studies have demonstrated the heterogeneity of the brain network, especially between the two cerebral hemispheres \cite{b17, b18, b19, b20}. For example, the right hemisphere has been shown to be dominant in responding to the environment and processing emotion \cite{b21, b22}, while the left hemisphere is dominant in language processing \cite{b23, b24}. 
Therefore, nodes located in different hemispheres should be considered to have different properties and traits. 
Furthermore, as key features of the human brain, hemispheric specialization, also known as brain asymmetry, is considered to be relevant for the early identification of neurological diseases \cite{b25, b26}. 
As a vivid example, a statistical analysis of the Alzheimer's Disease Neuroimaging Initiative (ADNI) dataset is presented in Fig.~\ref{fig:DTI_divided}. 
The DTI-based brain network edges are categorized into 3 types: left/right intrahemispheric edges and interhemispheric edges (Fig.~\ref{fig:DTI_divided}-a). 
Then, it is observed that the left and right intrahemispheric edges have similar strengths, and both have absolutely higher strengths than the interhemispheric edges (Fig.~\ref{fig:DTI_divided}-b). 
This phenomenon indicates that intra- and interhemispheric edges may exhibit different patterns in the human brain network. 
Thus, a homogeneous brain network is not sufficient to model the complex brain states and activity, especially in terms of the heterogeneity of intra- and inter-interactions between the two hemispheres.

Nevertheless, previous studies \cite{b70, b71, b72, DBLP:conf/miccai/YaoYSSL21} only referred to ``heterogeneous features'' in regard to the modality of brain neuroimaging, which deviates from the nature of heterogeneity in graph and brain network analysis.
In fact, brain regions are connected by different kinds of functional connections caused by hemispheric lateralization and other mechanisms. 
In other words, the brain network requires multitype edge and node modelling, regardless of how many modalities are used.  

Faced with the abovementioned issues and challenges, in this paper, we introduce the concept of a ``heterogeneous graph" into brain network analysis based on cerebral hemispheric heterogeneity. 
Specifically, we model the brain network as a heterogeneous graph with multiple types of nodes and edges under the hypothesis that left and right brain nodes operate with different rhythms and working patterns. Furthermore, we propose a heterogeneous graph neural network, namely, HebrainGNN, for multimodal heterogeneous brain network fusion learning. 
Our proposed model is considered capable of capturing rich and complex information regarding the heterogeneity between the two hemispheres. 
In addition, we propose a pretraining strategy designed for a heterogeneous brain network based on the contrastive learning method \cite{b27, b28}. 
This strategy helps us alleviate the overfitting problem due to the limited labelled sample size, which commonly occurs in brain-related datasets. 
Experimental results on two datasets show the superiority of the proposed model compared with state-of-the-art methods.

Our main contributions are summarized as follows: 
\begin{itemize}
    \item We propose a new perspective on modelling the brain network as a heterogeneous graph. We theoretically and practically present the inspiration that arises from the heterogeneity between the two hemispheres and the asymmetry of the human brain.
    \item We propose the HebrainGNN to integrate neuroimaging data acquired by fMRI and DTI. This model is able to encode heterogeneous brain networks and shows powerful graph representation learning capabilities.
    \item We also propose a self-supervised pretraining strategy based on a heterogeneous brain network, which helps alleviate the problems due to limited labelled data. 
\end{itemize}

\section{Related Works}
\subsection{Multimodal Brain Network Learning}
The goal of multimodal fusion learning based on brain networks is to utilize the strengths of each neuroimaging modality and build potential cross-modal feature spaces \cite{b6, b7, b15}. 
Unsupervised brain network learning methods usually attempt to find a common node embedding shared across all modalities \cite{b29, b30, b31, b32}. 
Such embeddings with multimodal fusion are considered to achieve better performance in downstream tasks such as disease prediction. 
In addition, there is a growing number of supervised methods, especially GNN-based approaches \cite{b14, b33, b34}. 
For example, Zhang et al. proposed deep multimodal brain networks (DMBNs) \cite{b34}. 
The DMBN model attempts to reconstruct functional connections (FCs) with structural connections (SCs) as its input. 
Then, the node embeddings generated by the GNN encoder are read out as graph-level representations for the supervised learning task. 
In some studies, population-level graph learning has also been conducted based on single-modal \cite{b35, b13} or multimodal neuroimaging datasets \cite{b37}. 
However, this kind of semisupervised learning framework is transductive and has difficulties in online inference. 
Furthermore, as we emphasized in the Introduction section, these methods neglect the nature of the brain connectivity itself, especially the heterogeneity of connections across the two cerebral hemispheres. 
Thus, an inductive GNN that considers different types of connectivity and works at the individual level for multimodal brain network learning is required.

\subsection{Heterogeneous Graph Representation Learning}
Traditional graph embedding methods such as DeepWalk \cite{b9}, graph convolutional networks (GCN) \cite{b38} and graph attention networks (GAT) \cite{b39} usually assume that graph-format data contain nodes and edges of the same type. 
This simplification obviously results in the loss of considerable information since a real-world graph is usually a heterogeneous information network (HIN) consisting of multiple types of nodes and edges \cite{b40, b41}. 
Abundant models have been proposed for representation learning based on heterogeneous graphs or knowledge graphs \cite{Kgat_KDD19, KBert_AAAI20} and have achieved ascendant progress \cite{b42, b43, b44, b45}.
For instance, Wang et al. proposed a heterogeneous graph attention network (HAN) that utilizes a two-level attention mechanism (i.e., the node level and the semantic level) to capture complex structures and rich semantics in heterogeneous graphs \cite{b42}.
However, current heterogeneous graph-based methods usually focus on local graph tasks (such as node classification and link prediction) and unweighted graphs. 
Few studies have attempted to extend heterogeneous GNNs to global graph tasks and weighted graphs, such as weighted brain network classification. 
\begin{figure*}[htbp]
\centerline{\includegraphics[scale=0.30]{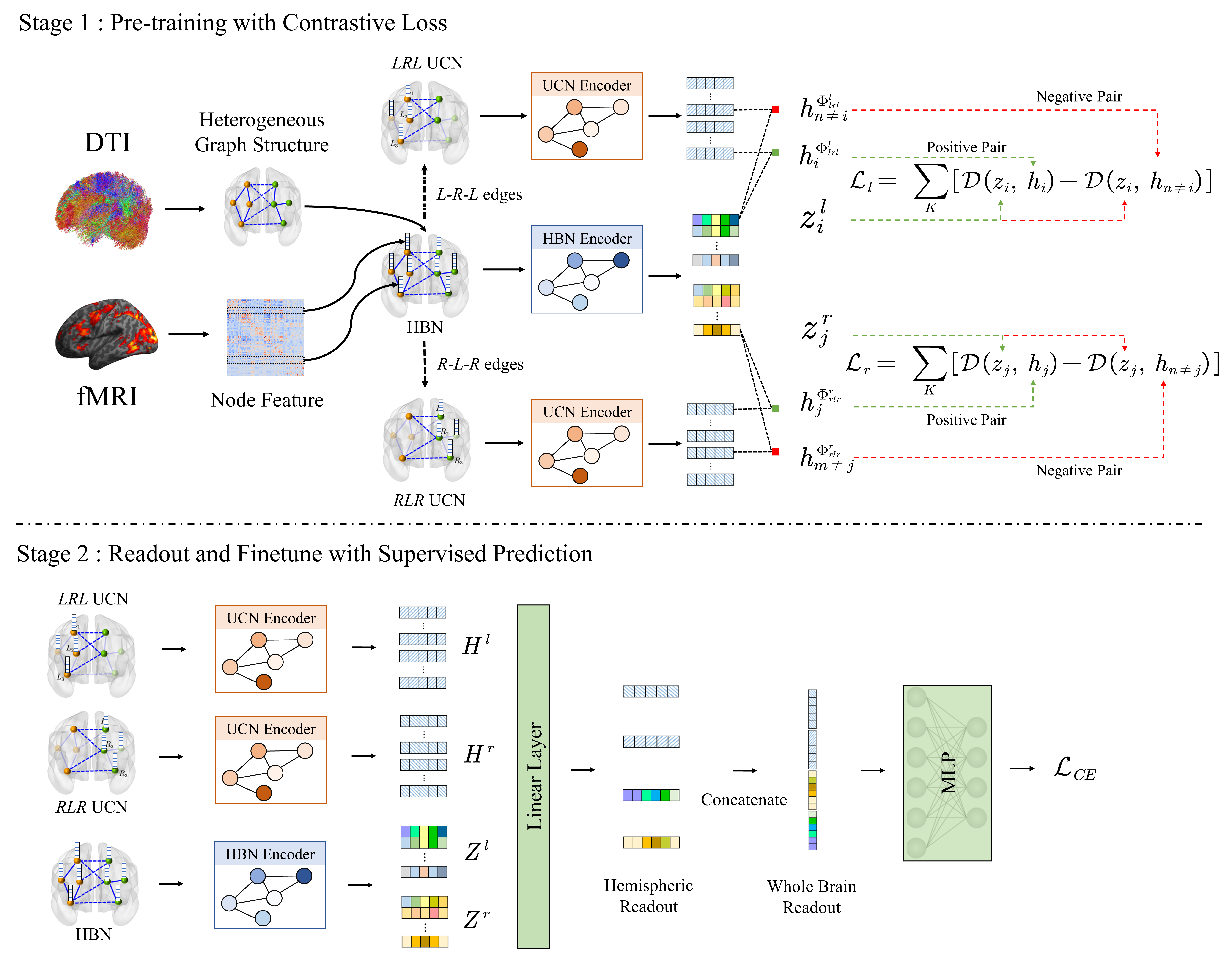}}
\caption{The overall framework of our proposed HebrainGNN model. Stage 1: Encoder part of heterogeneous brain network and self-supervised pre-training procedure. Stage 2: Supervised training procedure. HeteroEncoder and HomoEncoder are encoders for heterogeneous and homogeneous brain network, respectively (see section \ref{encoder part}).}
\label{fig:framework}
\end{figure*}

\subsection{Self-Supervised Learning (SSL) Based on the Contrastive Learning Method} The contrastive learning method is an important unsupervised approach for representation learning and has been widely used in areas such as natural language processing \cite{b46, b47}, computer vision \cite{b28, b48, b49} and data mining \cite{b27, b50, b51}. 
A score function is usually trained to classify the positive and negative pairs generated from the encoder.
For example, in the deep mutual infomax (DMI) model \cite{b28}, high-level and low-level representations from the same picture are considered positive pairs, while negative pairs are sampled from different pictures.
Such self-supervised methods enable deep learning models to generate high-quality representations in an unsupervised manner. Besides, Yan et al., proposed an GAN-based method to overcome the small data sample problem in brain imaging \cite{yan2021improving}. Chen et al. proposed a BrainNet model designed for SEEG brain network and adopted several self-supervised learning tasks for data denoising \cite{chen2022brainnet}.
In this work, we also propose a novel self-supervised pretraining method designed for heterogeneous brain networks. 
This method helps our model capture cross-hemispheric interactive information before the supervised signals are added and alleviates the problem of limited labelled data in medical image analysis.

\section{Methodology}
In this section, we introduce the proposed HebrainGNN model. 
The overall framework of HebrainGNN is presented in Fig.~\ref{fig:framework}. The DTI and fMRI brain networks are constructed as a graph structure and initial node feature matrix, respectively. In the pretraining procedure (Stage 1, introduced in section \ref{encoder part} and \ref{SSL part}) in detail, each node obtains two representations from heterogeneous and homogeneous graph encoders, and the model maximizes the mutual information between the two-view representations.

In the fine-tuning procedure (Stage 2, introduced in section \ref{readout part}) in detail, graph-level representations will be obtained through the two-view representations by using a linear layer. A multilayer perceptron (MLP) is used to obtain the prediction result with graph-level representations as input.

\subsection{Preliminary Concepts and Notations}
\label{notations}
\begin{figure}
\centerline{\includegraphics[width=\columnwidth]{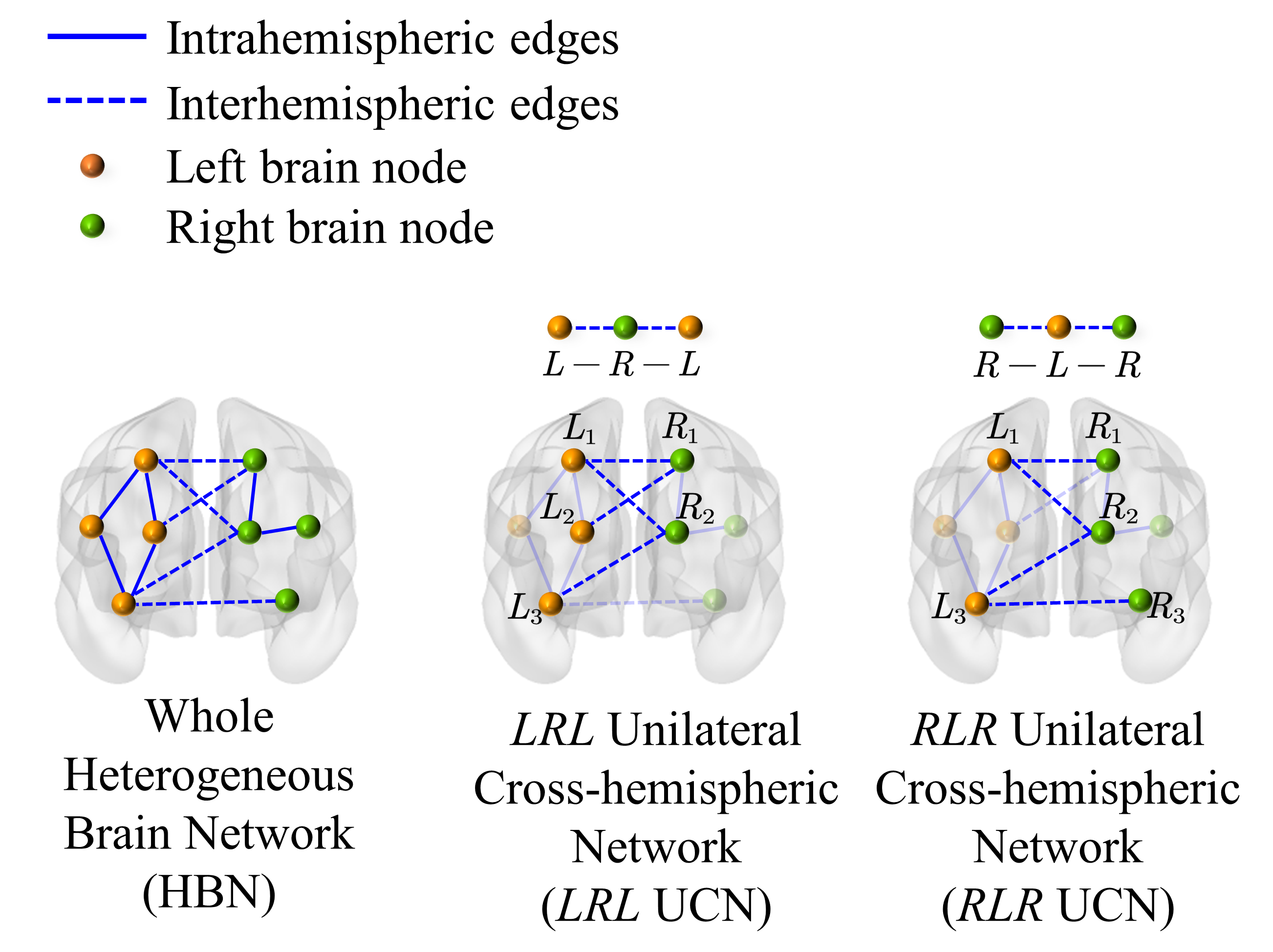}}
\caption{Illustration of the heterogeneous brain network (HBN) and unilateral cross-hemispheric network (UCN) used by HebrainGNN.}
\label{fig:Heterobrain_definition}
\end{figure}

\textbf{Heterogeneous Brain Network (HBN)}
An HBN in this paper is defined as a four-tuple $\mathcal{G}=\left( \mathcal{V}, \mathcal{E}, \mathcal{T}, \mathcal{R} \right) $, where $\mathcal{V}$ and $\mathcal{E}$ are the node set and edge set, respectively, while $\mathcal{T}$ and $\mathcal{R}$ denote the sets of node types and edge types, respectively. 
For a heterogeneous brain network, $|\mathcal{T}| +\,\,|\mathcal{R}| >\,\,2$. 
In this paper, we define two node types, which are left and right hemispheric nodes, and two edge types, namely, intrahemispheric and interhemispheric edges, in a brain network (Fig.~\ref{fig:Heterobrain_definition}). 
Furthermore, considering the strength distribution shown in Fig.~\ref{fig:DTI_divided} and for the purpose of simplification, we regard the left and right intrahemispheric edges as the same edge type. 

\textbf{Cross-Hemispheric Path} 
A multihop cross-hemispheric path $\Phi ^m$ is defined as a symmetric path that contains only cross-hemispheric edges. 
For example, the 3-hop path $L-R-L$ refers to a path that begins in the left hemisphere, connects to the right hemisphere and returns to the left hemisphere (Fig.~\ref{fig:Heterobrain_definition}).
We consider that cross-hemispheric paths in the brain network represent cross-hemispheric message interactions.
Considering that hemispheric lateralization or brain asymmetry is a fundamental characteristic of the human brain, we assume that such paths contain valuable information.

\textbf{Unilateral Cross-Hemispheric Network (UCN)}
A cross-hemispheric path represents an indirect neighbour relationship, which is different from a direct connection from the perspective of the whole brain network. 
Taking Fig.~\ref{fig:Heterobrain_definition} as an example, $L_2$ and $L_3$ are neighbours of $L_1$ through the paths $L_1-R_1-L_2$ and $L_1-R_2-L_3$. 
This kind of connection represents a different type of neural process compared with a direct connection such as $L_1-L_2$. 
Furthermore, latent neighbour relationships are uncovered by such paths. 
For example, $R_3$ can be regarded as a neighbour of $R_2$ via $R_3-L_3-R_2$, although there is no direct connection between them. 
For a path $\Phi ^m$, the neighbour set $\mathcal{N}^{\Phi ^m}$ for each node type $m$ can be obtained through such a neighbour relationship. 
Then, we construct a unilateral cross-hemispheric brain network $\mathcal{G}^{\Phi ^m}$ based on $\mathcal{N}^{\Phi ^m}$.
We highlight the differences between the HBN and the UCN here.
The HBN is a heterogeneous graph, while a UCN is a homogeneous graph that contains only single-hemispheric nodes. 
The HBN represents the 1st-order direct connections throughout the whole brain, while a UCN represents the 2nd-order cross-hemispheric connections between nodes in a single hemisphere.

\subsection{Brain Network Encoder in HebrainGNN}
In our proposed multimodal heterogeneous model, the DTI-based brain network is constructed as a graph structure, while the FCs from fMRI are represented as node features. 
There are two reasons for this modelling method. 
First, DTI data provide direct connections between brain regions, while FCs reflect the correlations of blood-oxygen-level-dependent (BOLD) signals between these regions. 
Thus, the DTI-based brain network is more suitable for reflecting the structural properties of the brain. 
Second, graph structures used in GNN models are commonly considered sparse, such as social networks and molecular networks. 
Compared with FC data, DTI-based brain networks are sparse and more consistent with this property when using a GNN.
\label{encoder part}

\textbf{HBN Encoder}. 
An HBN contains two types of nodes and two types of edges.
Therefore, a conventional homogeneous GNN model cannot directly encode HBN information. 
For each node, we consider the aggregation of messages from two edge patterns (i.e., intra- and interhemispheric edges) and finally generate the node representations accordingly. 
In addition, for each edge type, edge features such as connection strength are also taken into consideration.

Specifically, suppose the neighbour set $\mathcal{N}^{r}_i$, which denotes the set of neighbours connected to node $i$ by the edge of type $r$ (e.g., $\mathcal{N}^{r}_i$ can be either $\mathcal{N}^{L-L}_i$ or $\mathcal{N}^{L-R}_i$ for left-brain nodes), node $i$ receives messages sent from these neighbours through the edge-type-related projection matrix $W_r$. 
The representation of node $i$ is updated by aggregating the messages from all types of connected edges. 
The corresponding encoder is expressed as follows:

\begin{align}
 z^{l}_{i}
 =\sigma 
 \Big[ 
 \frac{1}{C(i)} 
 \big( 
 \sum_{r\in \mathcal{R}_i}
 &\sum_{j\in \mathcal{N}_{i}^{r}}
 {
 \frac{1}{\left|\mathcal{N}_{i}^{r} \right|}
 W_{r}^{\left( l-1 \right)} 
 f_{r}^{(l-1)}\left( e_{ij} \right) 
 \otimes 
 z_{j}^{\left( l-1 \right)} \big) 
 } 
 \Big]
 &  \notag \\
 +
 W_{o}^{\left( l-1 \right)}
 z_{i}^{\left( l-1 \right)}
 \label{eq1}
\end{align}
where $z_{i}^{\left( l \right)}$ is the hidden representation of node $i$ after graph convolution in layer $l$ and $C(i)$ is a normalization coefficient. 
We set $C(i)=\left| \mathcal{R}_i \right|$, where $\mathcal{R}_i$ is the set of edge types connected to node $i$. 
$\sigma \left( \cdot \right) $ is a nonlinear activation function such as the rectified linear unit (ReLU). 
$W_{r}^{\left( l \right)}$ and $W_{o}^{\left( l \right)}$ are the trainable parameters in layer $l$.

In the HBN Encoder, the edge features $e_{ij}$ are also taken into consideration. 
$f_{r}^{\left( l \right)}\left( \cdot \right) $ is a function that maps $e_{ij}$ to the same space as $z_{j}^{\left( l \right)}$, and $\otimes $ is the Hadamard product. 
In HebrainGNN, $\left| \mathcal{R}_i \right|=2$. 
The edge weights for the SCs are modelled as $e_{ij}$, and one linear layer is trained as the mapping function $f_{r}^{\left( l \right)}\left( \cdot \right) $.

\textbf{UCN Encoder}. 
A UCN is a graph for modelling cross-hemisphere path. 
We consider that the strength of interhemispheric edges has already been processed by the GNN encoder, and the strength of multihop connections is difficult to determine.

Therefore, in HebrainGNN, the UCN is constructed as a homogeneous and unweighted graph, and we adopt simple graph convolution (SGC) \cite{SGCN_ICML19} with a self-loop as the basic UCN encoder. 
This encoder preserves message passing in graph convolution with fewer model parameters, which can be expressed as:
\begin{equation}
H^{\Phi^{m}} = S_{\Phi^{m}}^{k} \cdot H \cdot W_{\Phi^{m}} + H \cdot W_{o} ,
\label{eq2}
\end{equation}
where $ S_{\Phi^{m}}$ is the symmetric Laplacian matrix for $A^{\Phi ^{m}}$ (i.e., $ S_{\Phi^{m}}=(D^{\Phi ^{m}})^{-\frac{1}{2}} A^{\Phi ^{m}} (D^{\Phi ^{m}})^{-\frac{1}{2}}$). $D^{\Phi ^{m}}$ is the degree matrix for $A^{\Phi ^{m}}$. $k$ is the order of SGC and $S_{\Phi^{m}}^{k}=\underbrace{S_{\Phi^{m}} \cdots S_{\Phi^{m}}}_{k}$. $W_o$ and $W_{\Phi ^{m}}$ are the trainable parameters for the cross-hemispheric path $\Phi ^{m}$. 

\subsection{Self-Supervised Pretraining Strategy}
\label{SSL part}
In our proposed pretraining procedure, our goal is to maximize the mutual information (MI) between $z_{i}$ (the output of the HBN encoder) and $h_{i}^{\Phi ^{m}}$ (the output of the UCN encoder) for each node $i$. 
This optimization procedure helps $z_{i}$ capture more information from $h_{i}^{\Phi ^{m}}$, thereby achieving the following benefits:
\begin{itemize}
    \item Enlarging the receptive field for graph convolution with a shallow model. The HBN represents only 1st-order connection relationships, while a UCN represents 2nd-order cross-hemispheric dependence relationships. Maximization of the MI helps both brain network encoders generate powerful representations $z_{i}$ with fewer graph convolutional layers and thereby decreases the risk of overfitting.
    \item Helping the UCN encoder capture cross-hemispheric semantics. A UCN characterizes cross-hemispheric neural processes, which are fundamentally different from direct connections in a single hemisphere. The optimization procedure provides an opportunity for the UCN Encoder to capture the features of such cross-hemispheric interactive messages.
    \item Dimension reduction before the introduction of supervised signals. The optimization procedure reduces the dimensionality of the node embeddings in the absence of supervised signals, which provides a good starting point for the optimization process of supervised learning. It also helps alleviate the problems that arise due to the limited availability of labelled data for common medical image analysis.
\end{itemize}

To estimate the MI between $z_{i}$ and $h_{i}^{\Phi ^{m}}$, the whole optimization procedure can be expressed as follows:
\begin{equation}
\hat{W}_{\mathcal{R}}, \hat{W}_{\Phi}=\underset{W_{\mathcal{R}}, W_{\Phi^{m}}}{\max}MI\left( Z,H_{}^{\Phi} \right),
\label{eq3}
\end{equation}
$W_{\mathcal{R}}$ and $W_{\Phi}$ are parameters for the HBN encoder and UCN encoder, respectively. 
Based on \cite{b28,b52,b53}, $MI$ can be estimated as:
\begin{align}
    & MI\left( Z,H^{\Phi ^{m}} \right) \propto  \max_{D(\cdot)} \notag \\
    & \Bigg(\mathbb{E}_{\left( Z,H^{\Phi ^{m}} \right) \sim p\left( Z \right) p\left( H^{\Phi ^{m}} \right)} \log \left( 1-\mathcal{D}\left( Z,H^{\Phi ^{m}} \right) \right) \notag \\
    & + \mathbb{E}_{\left( Z,H^{\Phi ^{m}} \right) \sim p\left( Z,H^{\Phi ^{m}} \right)} \log \left( \mathcal{D}\left( Z,H^{\Phi ^{m}} \right) \right) \Bigg) ,
\label{eq4}
\end{align}
where $p\left( Z,H^{\Phi ^{m}} \right)$ is the joint distribution for $Z$ and $H^{\Phi ^{m}}$, while $p\left( Z \right) p\left( H^{\Phi ^{m}} \right) $ is the product of their marginal distribution. 
The key is determining $p\left( Z,H^{\Phi ^{m}} \right)$ and $p\left( Z \right) p\left( H^{\Phi ^{m}} \right)$. 

Inspired by \cite{b27, b73}, $\mathbb{E}_{\left( Z,H^{\Phi ^{m}} \right) \sim p\left( Z,H^{\Phi ^{m}} \right)}$ is regarded as sampling $(Z,H^{\Phi ^{m}})$ from the same node (i.e., positive pair), while $\mathbb{E}_{\left( Z,H^{\Phi ^{m}} \right) \sim p\left( Z \right) p\left( H^{\Phi ^{m}} \right)}$ is sampling $(Z,H^{\Phi ^{m}})$ from different nodes (i.e., negative pair). 
The discriminator function $\mathcal{D}\left( \cdot \right) $ is trained to distinguish positive pairs and negative pairs. 
Therefore, for a specific node type $m$, the final optimization objective can be expressed as:
\begin{align}
    &\mathcal{L}_m(Z, H^{\Phi ^{m}}) = \frac{1}{C(m)} \sum_{i\in \mathcal{N}^{(m)}}{\sum_{\Phi^m_p\in \Phi^m}} \\ \notag
    &\Big(K \cdot \log \mathcal{D}( z_{i}, h_{i}^{\Phi^m_p} ) + \sum_{j\ne i}^K{\log ( 1-\mathcal{D}( z_{i}, h_{j}^{\Phi^m_p} ) )} \Big)
\label{eq5}
\end{align}
where $C(m)$ is the normalized coefficient. 
We set $C(m)=|\mathcal{N}^{(m)}|\cdot|\Phi ^m|$ in this paper. 
$K$ is the number of negative samplings, where a larger value of $K$ increases the difficulty of the contrastive learning procedure.

Considering that there are only two node types (i.e., $m \in \left\{ LN, RN \right\} $) and each node has only a single cross-hemispheric path (i.e., $|\Phi ^m|=1$), the whole optimization objective can be simplified as:
\begin{align}
&\mathcal{L}=\frac{1}{2}\sum_{m\in \left\{ LN, RN \right\}}^{}{\frac{1}{|\mathcal{N}^{(m)}|} \sum_{i\in \mathcal{N}^{(m)}}} \\ \notag
&{\left( K\cdot\log \mathcal{D}\left( z_{i}, h_{i}^{\Phi _{}^{m}} \right) + \sum_{j\ne i}^K{\log \left( 1-\mathcal{D}\left( z_{i}, h_{j}^{\Phi _{}^{m}} \right) \right)} \right)} .
\label{eq6}
\end{align}

We use a bilinear layer \cite{b27} with trainable parameter $W_D$ as the discriminator function:
\begin{equation}
\mathcal{D}\left( z_{i}, h_{i}^{\Phi _{}^{m}} \right)=\sigma \left( \left( z_{i} \right) ^{\mathrm{T}}W_Dh_{i}^{\Phi _{}^{m}} \right).
\label{eq7}
\end{equation}

\subsection{Readout and Prediction}
\label{readout part}
After the pretraining procedure, the output of the HBN encoder $Z$ and UCN encoder $H^{\Phi}$ are used for graph-level readout and supervised learning. 
For each subject $t$, every node has two representations $z_i$ and $h_i^{\Phi}$, and the final node representation is $H_t=\left( z_{1}, z_{2}, \cdots, z_{N}, h_{1}^{\Phi}, h_{2}^{\Phi}, \cdots, , h_{N}^{\Phi}\right) _{2N\times d}$ after pretraining, where $N$ is the number of predefined brain regions and $d$ is the hidden dimension. 
A linear layer is used to obtain the graph-level representation from $H_t$:
\begin{equation}
gh_t = WH_t + b,
\label{eq8}
\end{equation}
where $gh_t$ is the graph-level representation with shape $(1, 2N)$, and $W$ and $b$ are the trainable parameters in the linear layer. 
Then, an MLP is trained to derive the prediction for subject $t$.
This supervised procedure is expressed as:
\begin{equation}
\hat{y}_t=\mathrm{Softmax}{\mathrm{MLP}\left( gh_t \right)},
\label{eq9}\end{equation}
and cross-entropy is used to estimate the prediction loss:
\begin{equation}
\mathcal{L}_s=-\frac{1}{T}\sum_T{\left[ y_t\cdot \log \left( \hat{y}_t \right)+\left( 1-y_t \right) \cdot \log \left( 1-\hat{y}_t \right) \right]}.
\label{eq10}
\end{equation}

\section{Experiments}
\subsection{Datasets}
The experiments are performed on a private dataset and a well-known public dataset:
\begin{itemize}
    \item \textbf{Orthostatic Hypotension (OH) Dataset}: This dataset contains 224 subjects, including 147 patients suffering from OH and 77 healthy controls (HCs). All subjects were scanned to obtain DTI and fMRI images. We adopted the data augmentation strategy presented in \cite{b12} to balance the dataset. For each brain region in one subject, the mean time signal was extracted from a random 1/3 of the voxels instead of all voxels. We augmented the patient subjects 10 times and the HC subjects 20 times, resulting in a total of 1470 brain networks with OH and 1540 HC brain networks.
    \item \textbf{Alzheimer's Disease Neuroimaging Initiative (ADNI) dataset}\footnote{https://adni.loni.usc.edu/}: This dataset contains 512 subjects, including 250 cognitive normal (CN) subjects, 85 subjects with mild cognitive impairment (MCI), 84 subjects with early MCI (EMCI), 35 subjects with late MCI (LMCI), and 58 subjects with Alzheimer's disease (AD). For simplicity, subjects with MCI, ECMI, LMCI and AD were all regarded as patients in the experiment.
\end{itemize}
The preprocessing process and implementation details can be seen in appendix and our source code is at website\footnote{https://github.com/shigen97/HebrainGNN}.

\begin{table*}[htbp]
\caption{Disease prediction performance compared with baseline methods. The best and second-best results are shown in bold and underlined, respectively. Methods with "*" are heterogeneous graph-based methods.}
\begin{center}
\begin{tabular}{lllllllll}
\hline
Dataset    & \multicolumn{4}{c}{OH}                                                                    & \multicolumn{4}{c}{ADNI}                                                                  \\ \hline
Method     & Accuracy             & F1                   & AUC                  & Sensitivity          & Accuracy             & F1                   & AUC                  & Sensitivity          \\ \hline
GCN        & 63.80±05.79          & 61.88±05.90          & 62.70±07.49          & 61.51±05.79          & 73.40±00.99          & 74.67±00.97          & 72.28±02.06          & 75.66±01.96          \\
GAT        & 64.00±02.51          & 64.26±02.34          & 63.15±02.97          & 68.05±03.76          & 73.79±02.53          & 74.09±03.23          & 73.81±02.46          & 72.63±05.10          \\
CoRegSC    & 59.77±09.19          & 68.26±07.23          & 56.92±10.34          & 65.93±07.03          & 64.06±02.68          & 62.67±03.58          & 64.19±02.65          & 59.19±05.22          \\
GEHB       & 53.13±05.38          & 63.37±04.30          & 49.23±07.10          & 61.95±06.27          & 51.57±02.83          & 51.01±02.42          & 51.62±02.85          & 49.25±02.35          \\
ADB        & 65.86±06.47          & 66.96±06.56          & 65.50±06.58          & 71.88±09.46          & 74.95±02.70          & 73.90±03.09          &  \underline{77.43±03.61}    & 68.55±04.02          \\
MGCN       & 68.36±04.17          & 71.03±02.02          & 68.35±04.25          & 79.95±05.88          & 74.95±02.63          &  \underline{75.90±02.27}    & 74.90±02.68          &  \underline{76.04±03.11}          \\
DMBN       & 64.77±03.01          & 66.02±02.80          & 63.57±03.66          & 72.80±06.30          & 62.72±02.35          & 63.35±02.43          & 62.74±02.33          & 62.18±02.88          \\
Simple-HGN* & 67.43±02.96          & 69.14±03.08          & 71.06±03.72          & 73.07±04.12          &  \underline{75.73±00.87}    & 75.61±01.25          & 77.33±01.10          & 72.64±02.72          \\
HeCo*       &  \underline{72.00±04.37}    &  \underline{74.43±04.16}    &  \underline{75.68±02.62}    &  \underline{81.60±05.18}    & 72.23±01.45          & 72.00±01.66          & 75.49±00.57          & 68.92±02.41          \\
MeGNN*      & 67.07±03.20          & 69.09±03.63          & 66.11±03.84          & 73.93±05.89          & 72.43±00.48          & 72.12±01.21          & 72.93±01.66          & 68.93±03.91          \\
HebrainGNN(Ours)*  & \textbf{73.83±01.39} & \textbf{76.40±01.69} & \textbf{78.15±00.65} & \textbf{84.87±03.67} & \textbf{77.28±01.58} & \textbf{77.90±02.14} & \textbf{81.66±00.70} & \textbf{77.50±03.91} \\ \hline
\end{tabular}
\end{center}
\label{disese_prediction}
\end{table*}

\subsection{Baselines}
We compare the performance of disease prediction between the proposed HebrainGNN and existing baselines.
The baseline models include state-of-the-art brain network-based methods:
\begin{itemize}
\item \textbf{MVGE-HD}\cite{b31}: Multiview graph embedding with hub detection (MVGE-HD) is an unsupervised method. It considers hub detection in multiview graphs for dimensionality reduction, and experimental results show that the hub node detection mechanism can make two tasks interact such that each promotes the performance on the other.
\item \textbf{ADB}\cite{b33}: Attention-diffusion-bilinear (ADB) neural networks are supervised models. This method adopts an attention mechanism to generate representations from 1- and 2-hop brain networks. 
\item \textbf{MGCN}\cite{b74}: The multimodal graph convolutional network (MGCN) model introduced the idea of conducting graph convolution in rows and columns separately for brain network analysis.
\item \textbf{DMBN}\cite{b34}: Deep multimodal brain networks (DMBNs) proposed a new loss function. This model attempted to reconstruct FCs based on the node hidden representations generated from SCs.
\end{itemize}
The baseline models also include other competitive baselines:
\begin{itemize}
\item \textbf{GCN}\cite{b9} and \textbf{GAT}\cite{b39}. GCNs and GATs were originally designed for node classification tasks in social networks. A GCN simplifies spectral-domain graph convolution and performs spatial graph convolution via 1-hop neighbours. A GAT adopts a self-attention mechanism to determine the importance of neighbouring nodes and shows better performance than a corresponding GCN model.
\item \textbf{CoRegSC}\cite{b30}. Coregularized spectral clustering (CoRegSC) is an unsupervised method. It is a centroid-based coregularization algorithm that generates central embeddings for multiview data.
\item \textbf{Simple-HGN}\cite{simpleHGNN_KDD21}. This method simplifies the encoding procedure of a heterogeneous graph network (HGN). It embeds the complex relationships between nodes and incorporates the edge embeddings into the GAT model.
\item \textbf{HeCo}\cite{HeCo_KDD21}. This model conducts contrastive learning in two views of a heterogeneous graph. It was also originally designed for node classification tasks for a single type of node, and we implement it for left and right brain nodes in this work.
\item \textbf{MeGNN}\cite{MeGNN_KBS22}. The metapath extracted heterogeneous graph neural network (MeGNN) embeds different node types, and message importance between nodes depends on target and source node types. In addition, it utilizes a multichannel mechanism. With several parallel channels employed in the model, this method adopts a channel consistency regularization loss to increase the robustness.
\end{itemize}

We report four metrics (classification accuracy, F1 score, area under the receiver operating characteristic curve (AUC) and sensitivity) for evaluation. We also randomly divide the data into a training set and a test set with rates of 0.8 and 0.2, respectively. 
The experiment is repeated 5 times, and the mean performance is reported.

\subsection{Main Disease Prediction Results}
We compare our HebrainGNN with previous state-of-the-art methods based on the two datasets, as shown in Table \ref{disese_prediction}.
The results show that our proposed model achieves the best performance among all methods. 
For the OH dataset, the proposed HebrainGNN model achieves state-of-the-art performance (73.83\% accuracy, 76.40\% F1 score, 78.15 AUC and 84.87 sensitivity). 
For the ADNI dataset, the proposed model also obtains the best prediction scores, with 77.28\% accuracy, an F1 score of 77.90\%, an AUC of 81.66\% and a sensitivity of 77.50\%. 
Overall, the proposed model achieves approximately 1-2\% higher performance in terms of the four metrics than the best existing methods.

\begin{figure*}[htbp]
\centerline{\includegraphics[scale=0.38]{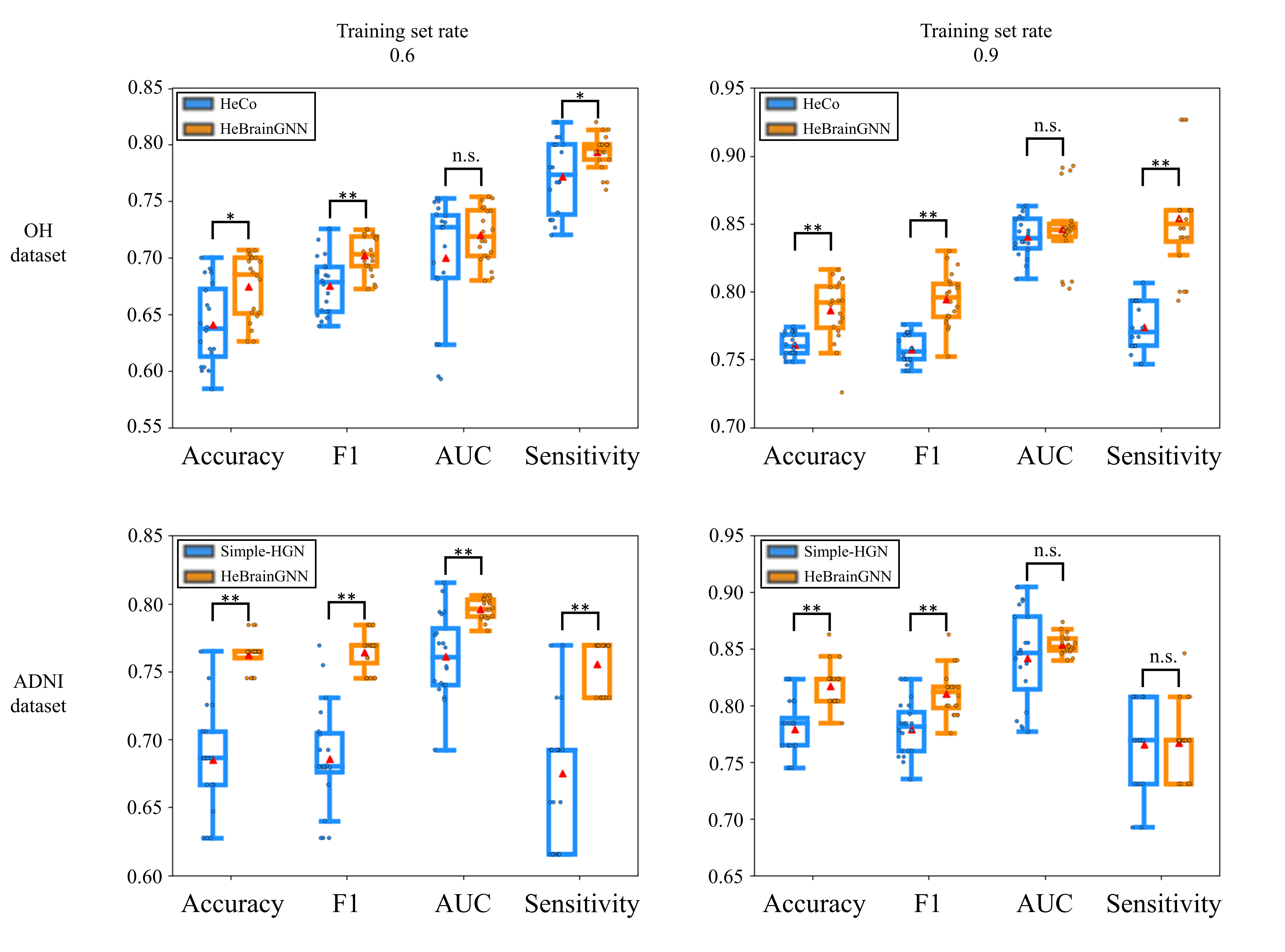}}
\caption{Statistical analysis of the prediction performance between our model and the best baseline in the two datasets (Simple-HGN for the OH dataset and HeCo for the ADNI dataset). We randomly sample the training set 20 times with rates of 0.9 and 0.6, and the same experiment is performed.}
\label{fig:ttest_results}
\end{figure*}

Furthermore, we present a statistical analysis to show the superiority of our model. 
Specifically, we randomly selected subsets of the training set at ratios of 0.6 or 0.9 and performed the disease prediction task. 
We repeated the same experiment 20 times and evaluated the performance difference between our proposed model and the second-best heterogeneous graph-based methods (HeCo and Simple-HGN for the OH and ADNI datasets, respectively). 
A paired $t$-test was also performed, and the results are shown in Fig.~\ref{fig:ttest_results}.
When the training rate is 0.9 (right column), there is a significant difference between our model and the compared method in terms of both the accuracy and F1 score. 
We observe no significant difference for the AUC metric in the two datasets, but the mean and median accuracies of our model are still higher than those of the compared method. 
When the training rate is 0.6 (left column), we observe a significant difference between our model and the compared model in terms of all the metrics, except the AUC metric in the OH dataset.
This observation indicates that our model derives better robustness, especially in the scenario of a small amount of data.

\begin{figure}[htbp]
\centerline{\includegraphics[width=\columnwidth]{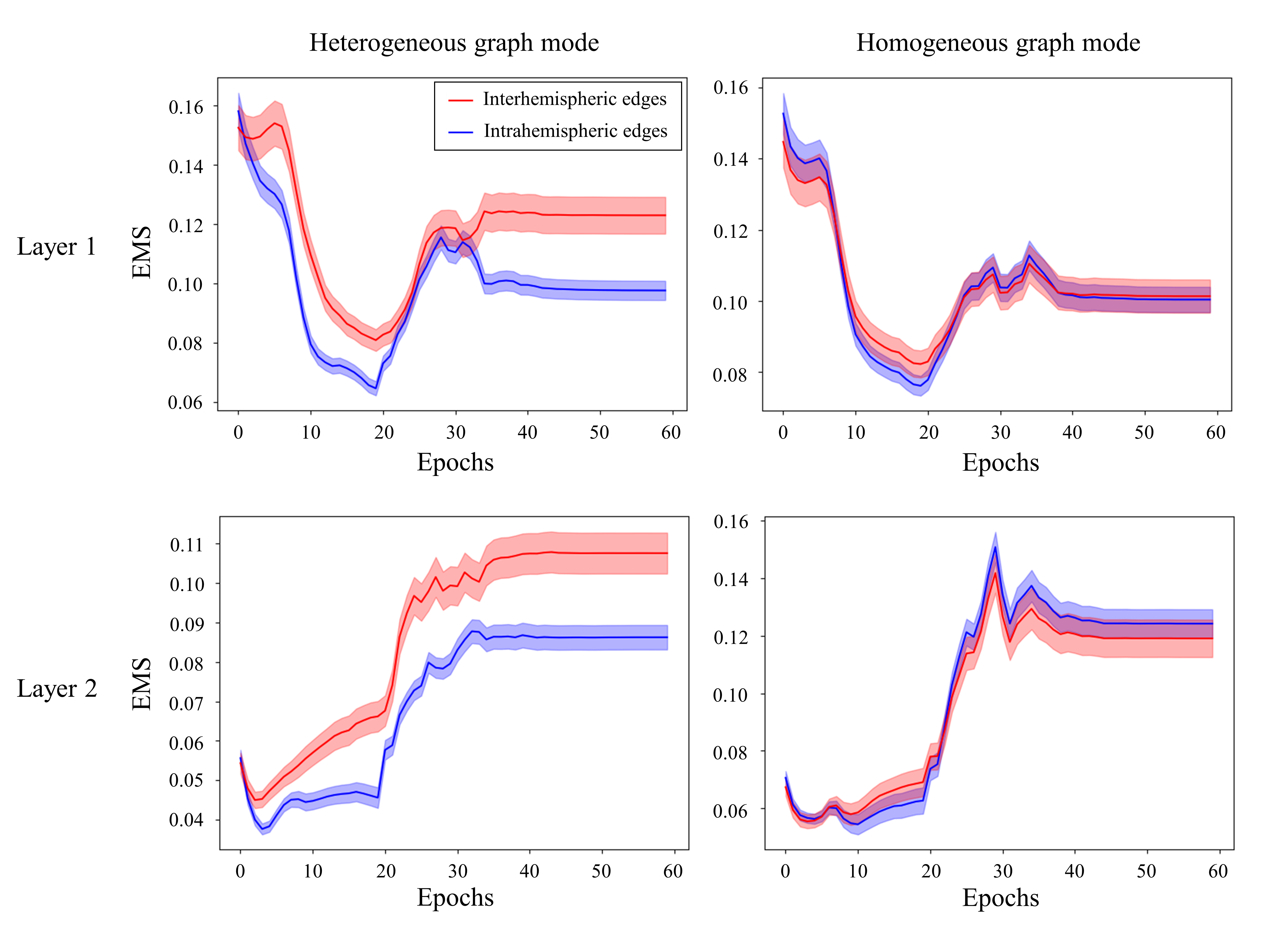}}
\caption{Variation in edge mapping scores in the HBN encoder (2 layers) for the two types of edges during training epochs.}
\label{fig:EMP_results}
\end{figure}

\subsection{Analysis of Hemispheric Heterogeneity}
We also investigate the influence of graph heterogeneity on our model. 
Compared with homogeneous graph-based models, the main difference in our model is that we define two different connection modes. 
That is, the heterogeneous nature of our model is mainly reflected in the edge mapping function $f_r(\cdot)$, including $f_{intra}(\cdot)$, $f_{inter}(\cdot)$, transformation matrix $W_{intra}$ and $W_{inter}$. 
Therefore, we try to investigate the patterns of hidden representations in the heterogeneous and homogeneous graph modes.

In the homogeneous graph mode, inter- and intrahemispheric edges are treated as edges of the same type, while the rest of the modelling is still the same as in the heterogeneous mode. 
We define an edge mapping score as follows:
\begin{equation}
EMS_r = \frac{1}{T}\sum_T\frac{1}{N}\sum_{i}{z^{r}_{i}}
\end{equation}
where $T$ is the total number of subjects and $N$ is the number of nodes in the brain network. 
$z^{r}_{i}$ is the node representation component for edge type $r$. 
$EMS_r$ represents the activation of edge type $r$ across the subjects. 
We take the ADNI dataset as an example and show the evolution of $EMS_{Inter}$ and $EMS_{Intra}$ during the training procedure in Fig.~\ref{fig:EMP_results}.

We observe that in the heterogeneous graph mode, $EMS_{Inter}$ almost maintains a higher value than $EMS_{Intra}$ during the training procedure in the two encoder layers. 
However, in the homogeneous graph mode, the patterns of $EMS_{Intra}$ and $EMS_{Inter}$ are different. 
$EMS_{Inter}$ is always similar to $EMS_{Intra}$ and shows no significant difference. 
From this phenomenon, we speculate that the heterogeneous graph mode enables much more activation of interhemispheric edges, and this mode assigns more importance to interhemispheric edges that the homogeneous mode may neglect because of their low connection strength.

To demonstrate this conjecture, we performed three additional experiments. 
We repeated the disease prediction experiment with a homogeneous graph model. 
In addition, we deleted the intrahemispheric edges and the interhemispheric edges separately to observe the performance changes in prediction. 
The results are given in Table \ref{abaltion_result}. 
We observe that with the homogeneous graph model, the prediction performance drops by approximately 5-10\%. 
More importantly, the model that considers only interhemispheric edges shows better performance than the model that considers only intrahemispheric edges. This finding also suggests that the interhemispheric edges play an important role in brain network analysis, but they may be neglected in homogeneous graph-based methods.

\begin{table*}[htbp]
\caption{Ablation study in the two datasets. 
HebrainGNN-homo mode denotes the model in homogeneous graph format. HebrainGNN-intra edges and HebrainGNN-inter edges denotes the graph data contain intra- or interhemispheric edges. HebrainGNN-SSL denotes the model contains only SSL procedure and the generated embeddings will be evaluated by logistical regression.  HebrainGNN-Scratch denotes the model trains from scratch without SSL procedure.}
\begin{center}
\begin{tabular}{lllllllll}
\hline
Dataset     & \multicolumn{4}{c}{OH}                                & \multicolumn{4}{c}{ADNI}                              \\ \hline
Method      & Accuracy    & F1          & AUC         & Sensitivity & Accuracy    & F1          & AUC         & Sensitivity \\ \hline
HebrainGNN-homo mode & 69.80±01.95&71.12±02.13&69.80±01.95&74.47±03.11&72.23±01.45&72.00±01.66&75.49±00.57&68.92±02.41 \\
HebrainGNN-intra edges & 61.33±01.50&63.30±02.07&63.25±01.40&66.80±03.44&68.74±01.67&66.14±02.89&70.93±01.30&59.19±04.97 \\
HebrainGNN-inter edges & 65.53±01.00&67.06±00.89&69.78±02.05&70.20±02.14&71.26±02.09&68.62±02.61&76.08±00.82&60.69±03.33 \\
HebrainGNN-SSL     & 69.90±01.37 & 69.38±01.95 & 71.85±01.13 & 65.91±03.70 & 65.57±02.14 & 69.29±02.06 & 73.99±01.85 & 77.73±02.98 \\
HebrainGNN-Scratch & 75.34±00.99 & 75.03±01.34 & 81.48±00.67 & 71.54±02.62 & 71.67±01.91 & 74.87±02.20 & 71.67±01.91 & 84.60±04.06 \\
HebrainGNN         & 77.28±01.58 & 77.90±02.14 & 81.66±00.70 & 77.50±03.91 & 73.83±01.39 & 76.40±01.69 & 78.15±00.65 & 84.87±03.67 \\ \hline
\end{tabular}
\end{center}
\label{abaltion_result}
\end{table*}

\subsection{Analysis of the Pretraining Strategy}
We further investigate the influence of the proposed self-supervised pretraining strategy in HebrainGNN. 
We show the results of our model without a pretraining strategy (HebrainGNN-Scratch) and without a fine-tuning procedure (HebrainGNN-SSL) in Table~\ref{abaltion_result}. 
When the model trains from scratch, the accuracy, F1 score, AUC and sensitivity metrics are all decreased by approximately 2\% for both datasets. 
The results demonstrate the effectiveness of the pretraining strategy for our proposed model.

Notably, the heterogeneity of the neuroimaging data may strongly influence the performance of SSL. 
The results in Table~\ref{abaltion_result} show that the SSL method without fine-tuning performs well on the OH dataset but poorly on the ADNI dataset. 
In fact, SSL attempts to learn the relationships contained within the data. 
Therefore, neuroimaging data collected from multicenter sites may be too heterogeneous to permit good model performance to be achieved in a self-supervised manner. 
This may also be one reason why the DMBN model achieves worse performance on the ADNI dataset. 
The DMBN model attempts to reconstruct the FCs with the SCs as input, and the different patterns generated due to the data heterogeneity may impede its performance.

\begin{figure}
\centerline{\includegraphics[width=\columnwidth]{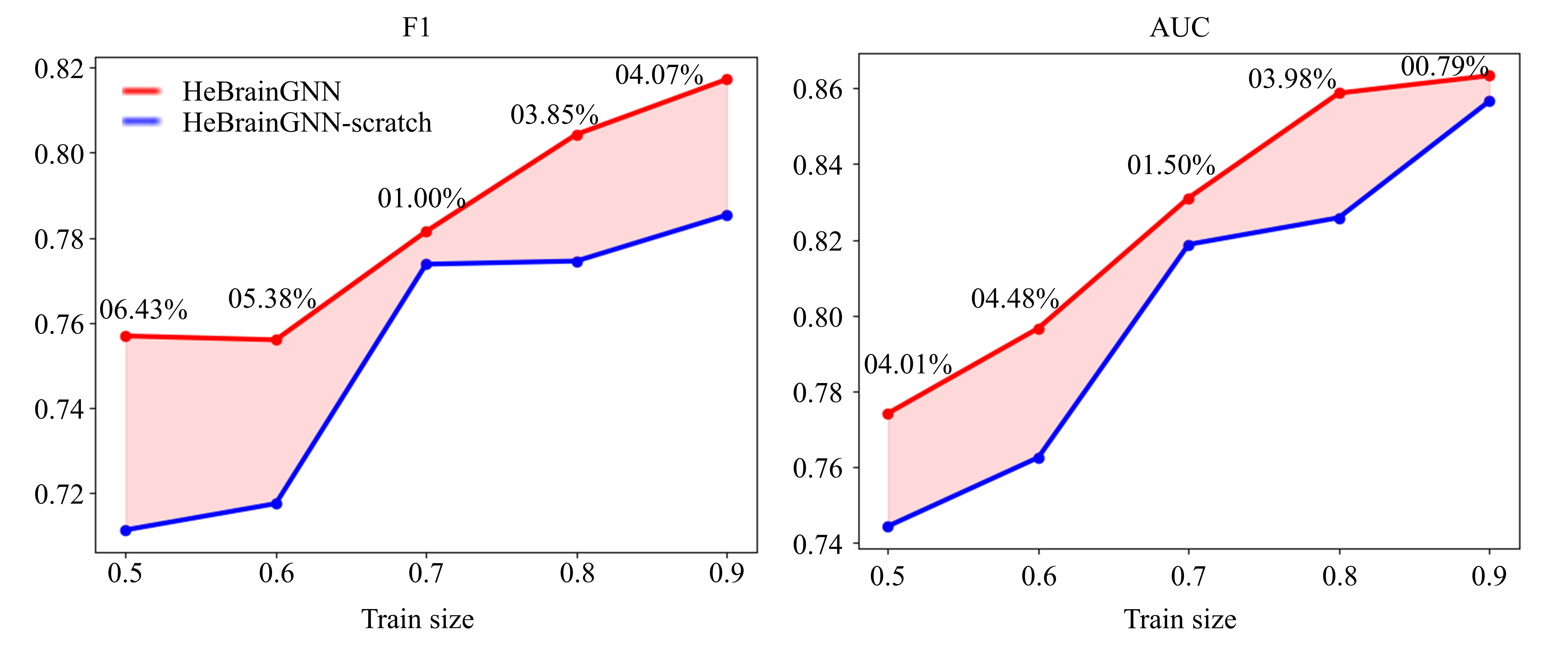}}
\caption{Variation of the effect of the pretraining strategy with training size. The percentage number represents the relative improvement rate of HebrainGNN compared to the model without the pretraining strategy. }
\label{fig:train_size_results}
\end{figure}

In addition, we investigate the influence of the training set size. We take the public ADNI dataset as an example here. 
We randomly selected a subset of the training set at a ratio r ($0.5 \sim 0.9$) and performed the same experiment by using our model with and without the pretraining procedure. 
We repeated the experiment 5 times and calculated the mean AUC and F1 score. 
The results are presented in Fig.~\ref{fig:train_size_results}. 
As the number of samples in the training set decreases, the performance declines, and the difference between our model with and without pretraining gradually increases. 
We also show the relative improvement rate in Fig.~\ref{fig:train_size_results}. 
These results demonstrate the superiority of our model on small labelled datasets and indicate that our proposed pretraining strategy alleviates the problems arising due to limited labelled samples in medical image analysis.

\subsection{Parameter Sensitivity Analysis}
\begin{figure}[htbp]
\centerline{\includegraphics[scale=0.25]{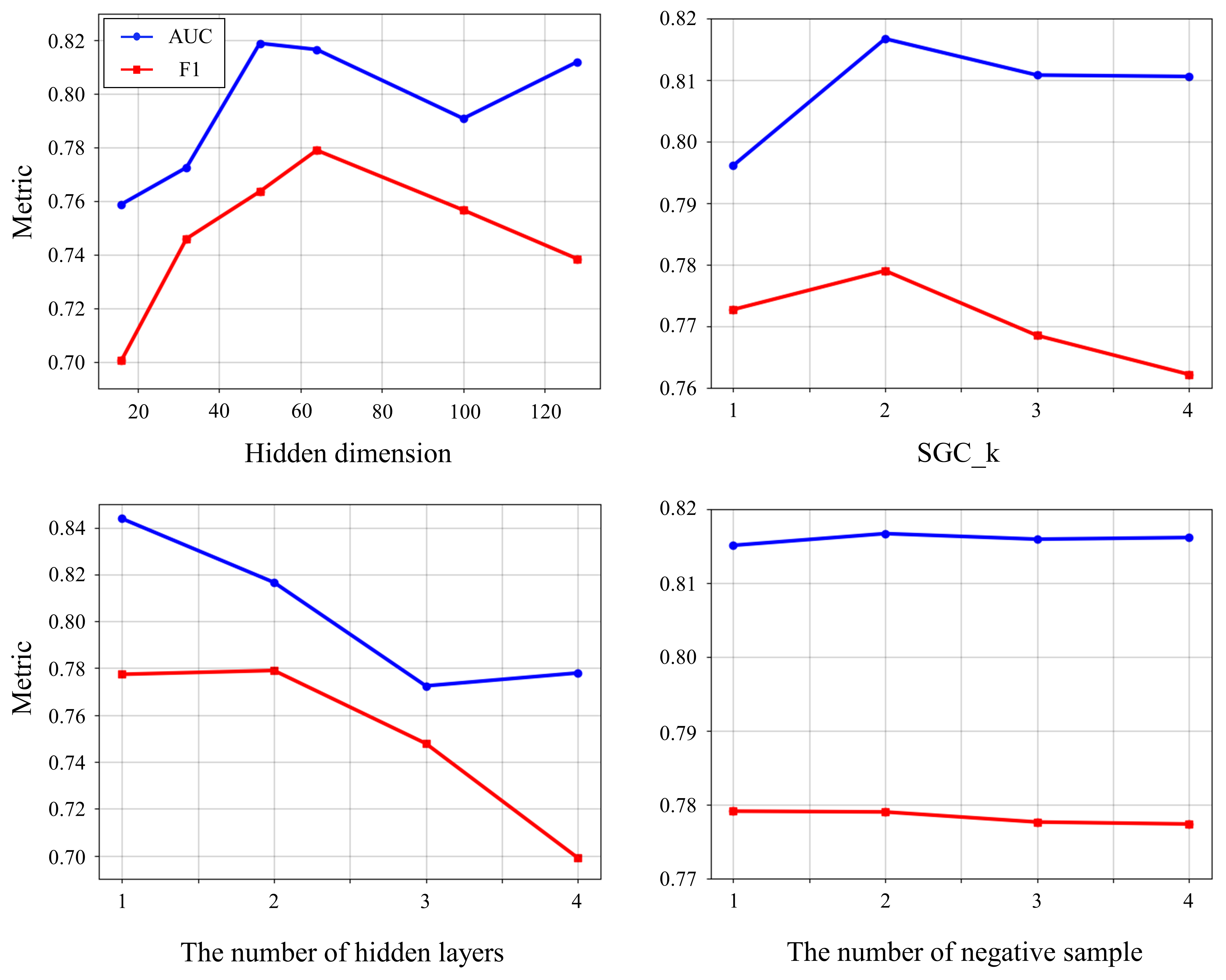}}
\caption{Sensitivity of the four main hyperparameters in HebrainGNN on the ADNI dataset.}
\label{fig:para_sensi}
\end{figure}

The sensitivity of the hyperparameters is also studied.
The four main hyperparameters include the hidden dimension of the nodes, the number of hidden layers, the order of SGC and the number of negative samples. 
The results are shown in Fig.~\ref{fig:para_sensi}.

It is observed that the number of negative samplings has a minor impact on the model performance. 
A small value of $K$ is sufficient for the dataset. 
However, compared with a negative sampling number, the hidden dimension of nodes and the order of SGCs have a relatively greater impact on model performance. 
With the increase in these two parameters, the performance first increases and then declines. 
This result indicates that a suitable hidden dimension and order of SGC are needed for the model. 
In addition, more hidden layers do not contribute any performance gains, and we empirically set the number of hidden layers to 2.

\section{Discussion and Limitations of the Study}

\begin{figure}
\centerline{\includegraphics[width=\columnwidth]{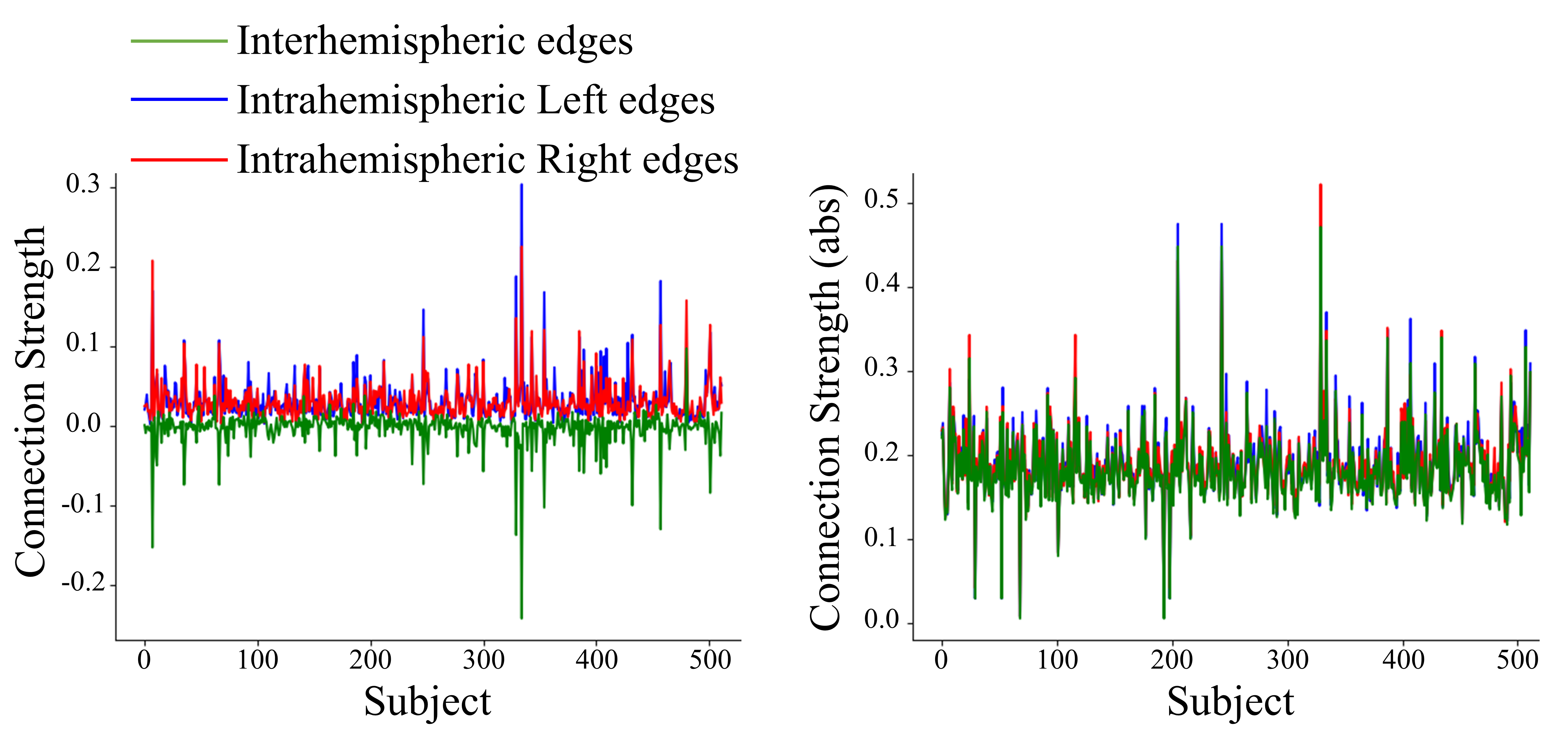}}
\caption{Mean strength of intra- and interhemispheric functional edges across subjects in the ADNI dataset. Because functional connections contain negative edges, we calculate the mean strength in two ways. One approach is to calculate the mean value for both positive and negative edges (left). The other way is to regard the absolute values of edges as the strength (right).}
\label{fig:fMRI_divided}
\end{figure}

Based on the investigation of HebrainGNN and its experimental results, we reassessed our study to facilitate the following discussion. 
First, the idea of HebrainGNN is inspired by the heterogeneity of intra- and interhemispheric connections in DTI-based brain networks. 
In fact, we also observe a similar phenomenon in fMRI brain networks (Fig. \ref{fig:fMRI_divided}, left column). 
In the ADNI dataset, intrahemispheric edges are stronger than interhemispheric edges in fMRI-based brain networks. 
This finding may reveal the assortativity of the DTI- and fMRI-based brain networks.
That is, if two brain nodes have a higher SC strength in the DTI-based network, they also tend to have a higher FC strength in the fMRI-based network. 
This finding may also show that hemispheric heterogeneity is a natural property of the brain that can be comprehensively characterized based on images acquired in multiple modalities instead of any specific modality.
Thus, we argue that the design of the characteristics of the data themselves is complementary to the design of the model. 

It is also notable that since fMRI-based brain networks originate from different DTI-based brain networks, they contain many negative value connections, representing a negative collaborative correlation between brain regions. 
We believe that the absolute values of negative edges may better represent their connection strength. %
Therefore, for the purpose of more precisely validating the functional hemispheric heterogeneity, we calculated the absolute values as the strength of the connections for all functional edges (Fig. \ref{fig:fMRI_divided}, right column).
Although the three types of edges seem to no longer have significant intuitive distinctions, statistical significance by paired $t$-test still exists between the intrahemispheric edges (both left and right) and the interhemispheric edges, with $p<0.05$.

Beyond the results, there are several limitations of the current study that may highlight directions for future studies.

\textbf{(1) More fine-scale modelling}. 
In this paper, we introduce the concept of heterogeneous graphs into brain network encoding and analysis. 
However, the modelling method used in the current study is still preliminary. 
The work of modelling heterogeneous brain networks with specific feature information for specific brain disease analysis is still required. 
For example, for cognitive-related diseases such as autism spectrum disorder, it may be necessary to construct node and edge types based on functional brain areas \cite{b57,b58}. 
For diseases that have a significant impact on the physical structure of the brain, such as Parkinson's disease, building a heterogeneous brain network in accordance with the relative positions of nodes in the brain is supposed to be more effective \cite{b59}.

\textbf{(2) More effective pretraining strategy}. 
The overfitting issue is often encountered in many small-sample medical image analyses, especially when the model becomes complex. 
The traditional transfer learning method (i.e. supervised pretraining on another dataset) may not work well because of the limited availability of labelled data. 
A more effective solution is to pretrain the model in a self-supervised or unsupervised manner, which requires only unlabelled samples. 
Therefore, pretraining with publicly accessible datasets, such as Parkinson's Progression Markers Initiative (PPMI) \cite{b60}, Autism Brain Imaging Data Exchange (ABIDE) \cite{b61,b62}, ADNI, and Human Connectome Project (HCP) \cite{b63}, shows bright prospects but still requires a unified SSL paradigm.

\textbf{(3) Dynamic temporal state analysis of the brain}. 
The human brain is a complicated dynamic system. 
However, the presented study still regards the brain network as a static graph.
Although some studies have explored the application of dynamic GNN models to neuroimaging data \cite{b64,b65,b66}, there have been few studies of dynamic heterogeneous brain networks to date. 
In addition, the temporal resolution of fMRI data is not sufficient to explore the transient dynamic changes in the brain. 
Thus, future studies may consider more multimodal sources with higher temporal resolution, such as electroencephalography (EEG) and magnetoencephalography (MEG), for characterizing dynamic states in heterogeneous brain networks, especially for cognitive and emotional recognition tasks \cite{b67,b68,b69}.

\section{Conclusion}
In this paper, we propose a novel HebrainGNN model for encoding heterogeneous brain networks. 
This model is inspired by the phenomena of hemispheric lateralization and brain asymmetry. 
We innovatively propose that intrahemispheric and interhemispheric edges exhibit different patterns and properties in brain network analysis. 
The HebrainGNN takes advantage of fusing multimodal neuroimaging data, and it shows superiority compared with other state-of-the-art methods. 
An analysis of edge mapping scores shows that our model attaches more importance to interhemispheric edges and thereby achieves better performance than other models. 
In addition, we propose a novel self-supervised pretraining strategy designed for heterogeneous brain networks, which is considered to alleviate the problems associated with limited training samples in medical imaging analysis.

\section*{Acknowledgements}
The data used in the preparation of this article were obtained from the Alzheimer's Disease Neuroimaging Initiative (ADNI) database (\underline{adni.loni.usc.edu}). The ADNI was launched in 2003 as a public--private partnership led by Principal Investigator Michael W. Weiner, MD. The primary goal of the ADNI has been to test whether serial magnetic resonance imaging (MRI), positron emission tomography (PET), other biological markers, and clinical and neuropsychological assessment can be combined to measure the progression of mild cognitive impairment (MCI) and early Alzheimer's disease (AD).

\bibliographystyle{IEEEtran}
\bibliography{main}

\IEEEdisplaynontitleabstractindextext

\IEEEpeerreviewmaketitle

\clearpage
\section*{Appendix}
\begin{figure}[ht]
\centerline{\includegraphics[width=\columnwidth]{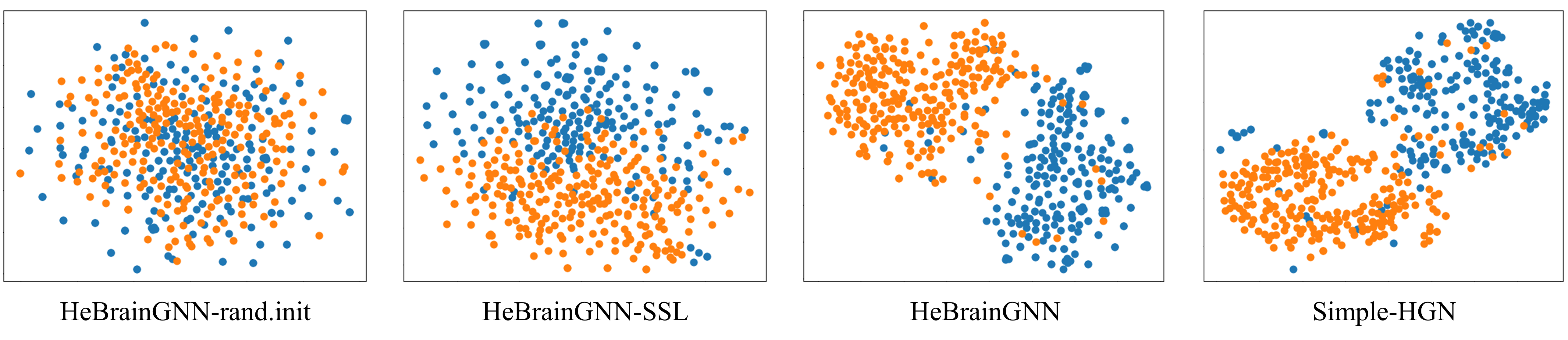}}
\caption{The t-SNE visualization of graph-level embeddings generated from the model. The color of the points represents the ground truth.}
\label{fig:visulization}
\end{figure}
\subsection*{Preprocessing and Experimental Setup}
For each subject in the OH dataset, the fMRI imaging data were acquired using a 3T GE scanner (
Signa HDxt, GE Healthcare
) with an 8-channel head coil. 
Resting-state data were acquired using a gradient-echo echo planar imaging (EPI) sequence. 
For resting-state fMRI scanning, the following parameters were used: 
repetition time (TR) = \SI{2000}{ms}; echo time (TE) = \SI{30}{ms}; field of view (FOV) = $240\times240$ \SI{}{mm^2}; acquisition matrix = $64\times64$; FA = $90\si{\degree}$; slice thickness = 4.4 mm; and 32 transverse slices. Each resting-state scan lasted 8 min and 20 s (250 vol). DTI data were collected using an EPI sequence with the following parameters: b-value = \SI{1000}{s/mm^2}, flip angle = $90\si{\degree}$, TR= \SI{13500}{ms}, TE = \SI{87.4}{ms}, FOV = $240\times240$ \SI{}{mm^2}, acquisition matrix = $240\times240$, and 50 contiguous slices, resulting in voxel dimensions of $1\times1\times3$ \SI{}{mm^3}. 
A total of 20 independent and noncollinear diffusion encoding directions and one additional image with no diffusion weighting (b = $0$) were acquired. 
High-resolution T1-weighted structural images were acquired in the coronal view with a slice thickness of 1 mm without a gap, flip angle = $15\si{\degree}$, TR = \SI{6.1}{ms}, TE = \SI{3.2}{ms}, and FOV = $240\times240$ \SI{}{mm^2} (matrix = $240\times240\times146$).

For the fMRI data, the graph theoretical network analysis (GRETNA) software toolbox was used \cite{b54}. 
The procedure consists of removing the first several volumes of images to ensure magnetization equilibrium, performing slice timing correction with the first slice, performing head-motion estimation and correction with a 0.01--0.10 Hz frequency bandpass, and coregistering the first scan of the fMRI time series with the T1-weighted images.
The preprocessed results were then normalized to the Montreal Neurological Institute (MNI) template space and spatially smoothed with a Gaussian kernel. 
To define the nodes of the brain network, the Brainnetome Atlas (246 regions) \cite{b55} was used for both datasets. 
The mean time series for each brain region was obtained by averaging the BOLD time series over the voxels. 
The edges of the fMRI brain networks were computed via Z-transformation of the values of the Pearson correlation coefficients.

For the DTI data, the deterministic tractography method was applied using the DSI Studio toolbox \footnote{http://dsi-studio.labsolver.org}. 
A deterministic fibre tracking algorithm \cite{b56} was used for fibre tracking after the diffusion tensor was calculated. 
The angular threshold was 70 degrees, and the step size was 1 mm. 
Tracks with lengths shorter than 20 mm or longer than 180 mm were discarded. 
The number of fibres connected to each of the remaining regions was calculated separately to define the edge weights between regions in the DTI brain network.

For the OH dataset, our HBN encoder contains 1 hidden layer, and the order $k$ of the UCN encoder is 2. The number of hidden dimensions is 64. 
The batch size is 128, and the dropout rate is 0.75. 
The learning rate is $1e-4$, and the L2 regularization value is 1e-5. 
The MLP for prediction contains 1 linear layer. 
The model is pretrained for 10 epochs with $K=1$ and then subjected to supervised training for 40 epochs (a total of 50 epochs). 

For the ADNI dataset, the HBN encoder contains 2 hidden layers, and the order $k$ of the UCN encoder is 2. 
The number of hidden dimensions is still 64. 
The batch size is 128, and the dropout rate is 0.7. 
The learning rates are 1e-4 and 2.5e-4 for pretraining and supervised training, respectively. 
The L2 regularization value is $1e-5$. 
The model is pretrained for 20 epochs with $K=2$ and then subjected to supervised training for 40 epochs (a total of 60 epochs). 
The learning rate decays by 75\% every 5 epochs after 35 epochs. 

\subsection*{Visualization}
We visualize the embeddings in the ADNI dataset generated by the proposed model and baseline model Simple-HGN to support an intuitive evaluation. 
The results are shown in Fig.~\ref{fig:visulization}. 
HebrainGNN-rand. init refers to the HebrainGNN model without a training procedure (the parameters are randomly initialized), and the corresponding embeddings are blurred across subjects. 
After the pretraining procedure, the embeddings of our model have a certain degree of distinguishability, although the boundaries are still blurred (HebrainGNN-SSL). 
The embeddings of the Simple-HGN method show a much better distinguishability. 
After the supervised fine-tuning procedure, the embeddings of our method show relatively clear boundaries compared with Simple-HGN.

\end{document}